\documentclass[conference]{IEEEtran}
\IEEEoverridecommandlockouts
\usepackage{cite}
\usepackage{amsmath,amssymb,amsfonts}
\usepackage{algpseudocode}
\usepackage{algorithm}
\usepackage{graphicx}
\usepackage{textcomp}
\usepackage{xcolor}
\usepackage{booktabs}
\def\BibTeX{{\rm B\kern-.05em{\sc i\kern-.025em b}\kern-.08em
    T\kern-.1667em\lower.7ex\hbox{E}\kern-.125emX}}
\begin{document}

\title{Proof of AutoML: SDN based Secure Energy Trading with Blockchain in Disaster Case\\}

\author{Salih Toprak\IEEEauthorrefmark{1}, Müge Erel-Özçevik\IEEEauthorrefmark{1}\\ 

\IEEEauthorblockA{\IEEEauthorrefmark{1}Department of Software Engineering, Manisa Celal Bayar University, Manisa, Turkey \\}

\IEEEauthorblockA{Emails: 222802069@ogr.cbu.edu.tr, muge.ozcevik@cbu.edu.tr}
}

\maketitle

\markboth{Accepted by 7th International Conference on Blockchain Computing and Applications (BCCA 2025), ©2025 IEEE}{}%

\begin{abstract}
In disaster scenarios where conventional energy infrastructure is compromised, secure and traceable energy trading between solar-powered households and mobile charging units becomes a necessity. To ensure the integrity of such transactions over a blockchain network, robust and unpredictable nonce generation is vital. This study proposes an SDN-enabled architecture where machine learning regressors are leveraged not for their accuracy, but for their potential to generate randomized values suitable as nonce candidates. Therefore, it is newly called Proof of AutoML. Here, SDN allows flexible control over data flows and energy routing policies even in fragmented or degraded networks, ensuring adaptive response during emergencies. Using a 9000-sample dataset, we evaluate five AutoML-selected regression models—Gradient Boosting, LightGBM, Random Forest, Extra Trees, and K-Nearest Neighbors—not by their prediction accuracy, but by their ability to produce diverse and non-deterministic outputs across shuffled data inputs. Randomness analysis reveals that Random Forest and Extra Trees regressors exhibit complete dependency on randomness, whereas Gradient Boosting, K-Nearest Neighbors and LightGBM show strong but slightly lower randomness scores (97.6\%, 98.8\% and 99.9\%, respectively). These findings highlight that certain machine learning models, particularly tree-based ensembles, may serve as effective and lightweight nonce generators within blockchain-secured, SDN-based energy trading infrastructures resilient to disaster conditions.
\end{abstract}

\vspace{-1em}
\section{Introduction}
In disaster scenarios, the collapse of traditional energy infrastructure and limited access to centralized communication networks create critical challenges for energy distribution and coordination. The emergence of decentralized energy resources, such as household-level solar panels and batteries, offers a promising solution for local energy sharing. However, to enable secure and transparent transactions among prosumers and consumers, a robust digital infrastructure is required.

Blockchain technology has emerged as a trusted mechanism for ensuring data integrity and transaction security in decentralized environments. Its consensus protocols rely on nonce values that must be unpredictable and unique to prevent hash collisions and ensure tamper-proof block generation. Traditionally, nonce generation has depended on hardware-based or pseudo-random number generators. However, these approaches may not perform reliably in resource-constrained or compromised environments, especially when the computational burden or entropy sources are limited.

Software-Defined Networking (SDN) offers centralized control and programmability over the underlying network infrastructure, which is essential in disaster settings where network topology may rapidly change \cite{b14}. The integration of SDN with blockchain-enabled energy trading provides a flexible and secure architecture for coordinating distributed energy exchanges and communication paths.

In this context, we explore the use of machine learning-based regression models as alternative nonce generators by leveraging their prediction variability. This shift aims to provide a lightweight yet effective approach to nonce generation by training models on throughput-related network features, thus bridging communication-layer intelligence with blockchain-layer security. Our motivation is rooted in the need for scalable, autonomous, and resilient systems that can operate under uncertain and dynamic disaster conditions.
\vspace{-0.5em}
\subsection{Related Work and Literature Review}
Blockchain is a distributed ledger technology that guarantees immutability, transparency, and decentralization by chaining data blocks via cryptographic hashes. Each block contains the hash of its predecessor, a timestamp, and a set of transactions; altering any block would break the chain’s integrity, making unauthorized changes easily detectable \cite{b1}. Early consensus mechanisms such as Proof-of-Work (PoW) ensure security through computational difficulty but incur high energy costs. More recent schemes like Proof-of-Stake (PoS) reduce energy consumption by requiring validators to lock up stakes rather than solve puzzles \cite{b1}, \cite{b2}.

The inherent security and auditability of blockchain make it attractive for critical infrastructures and emergency management. Costa et al. provide a comprehensive survey of how blockchain can support smart-city emergency response systems—ensuring reliable data sharing among heterogeneous IoT devices during crises \cite{b3}. In the context of disaster-resilient communications, Karaman et al. examine the integration of space-air-ground-sea networks powered by renewable energy sources after the 2023 Türkiye earthquake, highlighting blockchain-enabled peer-to-peer energy trade as a means to bolster network survivability \cite{b4}. Similarly, Zhou et al. propose resilience-oriented hardening and expansion planning for power transmission systems under hurricane impacts, recommending blockchain for secure coordination among distributed grid elements \cite{b5}.

In energy systems, blockchain facilitates trustworthy, automated transactions via smart contracts and decentralized marketplaces. Xu et al. develop a blockchain-based dispatching approach to balance high levels of renewable penetration, achieving transparent settlement and tamper-proof logging of energy flows \cite{b6}. Boumaiza and Sanfilippo introduce a testing framework for microgrid energy-trading applications, demonstrating how smart contracts can streamline peer-to-peer energy exchanges within local communities \cite{b7}. Kumari et al. further explore peer-to-peer transactive energy management in smart grids, showing that blockchain can enforce fair pricing and settlement without a central intermediary \cite{b8}.

Beyond basic energy trading, researchers have combined blockchain with advanced control and digital-twin techniques. Yu et al. integrate predictive secure control protocols with blockchain to defend networked energy systems against cyber threats \cite{b9}. Jamil et al. propose a digital twin–driven architecture for AIoT-based energy service provision, where blockchain ensures data integrity and trust in optimal energy trading among nanogrids \cite{b10, b15}. Ma et al. analyze the bibliometric landscape of blockchain-digital twin convergence, identifying key trends and research gaps in secure, real-time system modeling \cite{b11}. Most recently, Naeini et al. present PINN-DT, a hybrid physics-informed neural network framework secured by blockchain for optimizing energy consumption in smart buildings—illustrating the synergy between emerging AI techniques and dlt \cite{b12}.

\subsection{Contributions}

Therefore, we've proposed Proof of AutoML based energy trading for disaster scenarios. The main contributions of this paper are as follows:

\begin{itemize}
    \item We design a secure energy sharing and transaction system for disaster scenarios based on SDN and Blockchain technologies. 
    \item We propose the use of AutoML regression models for throughput prediction and evaluate their randomness capability for Blockchain nonce generation.
    \item We benchmark five models in AutoML — Gradient Boosting, LightGBM, Random Forest, Extra Trees, and K-Nearest Neighbors — in terms of both standard performance metrics (MAE, RMSE, R\textsuperscript{2}) and output randomness behavior.
    \item We provide a randomness rate analysis and identify which models are suitable for secure and diverse nonce generation in a Blockchain system for disaster case.
\end{itemize}

\section{Proposed Network Architecture}
\begin{figure*}
\centerline{\includegraphics[width=0.6\textwidth]{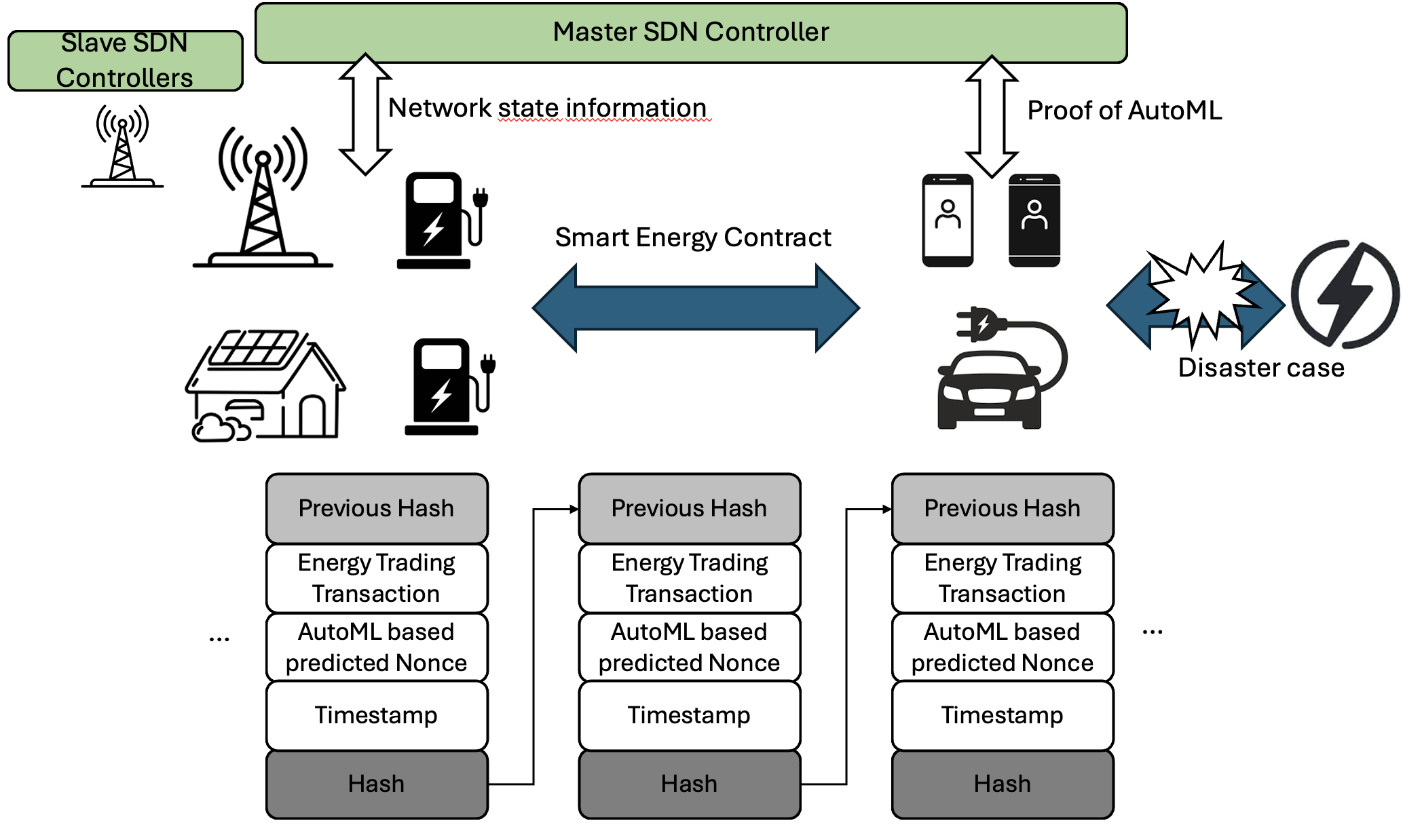}}
\vspace{-1em}
\caption{Proposed Proof of AutoML architecture for secure energy trading in disaster scenarios.}
\label{na_arch}
\vspace{-2em}
\end{figure*}
In Fig.\ref{na_arch}, the proposed network architecture for Proof of AutoML for secure energy trading is shown below. In the event of a disaster, users may not be supplied by the general energy center for a while. During this period, electric vehicles, users' phones, tablets, computers, servers, etc., would need to be recharged for optimal user experience. In that case, houses and base stations powered by solar panels and undamaged by the disaster are used as a power supply. However, the service fee for them may differ in a non-disaster case. Therefore, energy trading fees are determined by smart contracts signed by producers and consumers. Each energy trading request from users is signed by a smart energy contract and saved in a blockchain with the help of SDN controllers. SDN controllers have an AutoML model that produces a unique nonce value using the predictive characteristics of machine learning. There are many slave SDN controllers and a centralized master SDN controller. They communicate each other via OpenFlow protocol\cite{ONF} and each slave SDN controller (base stations) also configured by centralized SDN controller. In any communication failure with master SDN controller, energy trading blockchain can be also processed by slave SDN controllers. Thanks to this hybrid SDN architecture, energy trading is not affected and is stored in the blockchain. This stored energy can be used later in case of communication or energy supply failures in a disaster.
\section{Proposed System Architecture}
Natural disasters such as earthquakes or floods can severely disrupt traditional energy and communication infrastructures, rendering centralized energy management and data routing inoperative. In such scenarios, localized energy production units (e.g., solar panels) can play a vital role in sustaining essential services. However, enabling secure, decentralized, and efficient energy exchange under such constrained conditions presents significant challenges.
To address this, we propose an SDN-assisted blockchain architecture that enables peer-to-peer (P2P) energy trading between distributed nodes in post-disaster environments. The system integrates programmable SDN infrastructure for dynamic communication control, a blockchain ledger for secure transaction management, and AutoML-powered predictive intelligence to streamline the block mining process.

\begin{figure}
\centerline{\includegraphics[scale=0.7]{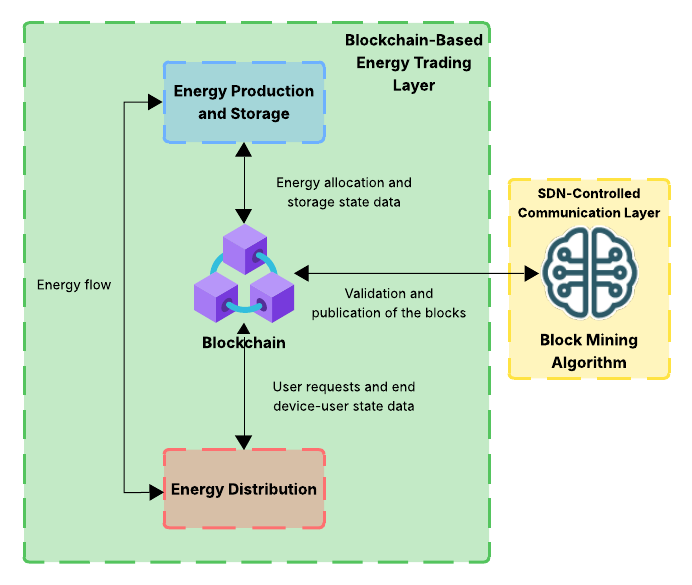}}
\vspace{-1em}
\caption{Proposed SDN-assisted blockchain architecture.}
\label{sys_arch}
\vspace{-1.5em}
\end{figure}

As illustrated in Fig.~\ref{sys_arch}, the architecture consists of two functional layers: (i) \textit{SDN-Controlled Communication Layer}, responsible for dynamic routing and device coordination; and (ii) \textit{Blockchain-Based Energy Trading Layer}, which oversees transaction validation and ledger maintenance. This modular design allows flexible deployment in partially connected environments with intermittent connectivity.

\subsection{SDN-Controlled Communication Layer}
Software-Defined Networking enables programmable control over data flows, making it well-suited for environments where communication paths are degraded. The SDN controller manages network state, routes energy-related transactions, and supervises the block mining process. Its ability to adapt to topology changes ensures energy packets and metadata reach their destinations with minimal delay, supporting real-time coordination among nodes.

\subsection{Blockchain-Based Energy Trading Layer}
Blockchain serves as the core trust mechanism, maintaining a tamper-proof ledger of energy transactions without requiring centralized validation. The consensus mechanism is integrated with the SDN controller to streamline validation. Features such as local caching and delayed synchronization are incorporated to ensure ledger consistency during temporary disconnections. Smart contracts may also be deployed to prioritize energy delivery for critical services.
\vspace{-0.5em}
\subsection{Integration with Predictive Intelligence}
To improve efficiency and reduce mining overhead, the block mining algorithm leverages AutoML-based regression models to generate pseudo-random nonce values derived from current network parameters. Instead of brute-force search, this approach enables lightweight nonce selection while ensuring sufficient variability. These models are deployed within the SDN control layer and are trained to balance randomness and computational simplicity—particularly beneficial for constrained, low-power devices common in disaster zones.
\vspace{-0.5em}
\section{Block Mining and Methodology}
\begin{figure}
\centerline{\includegraphics[scale=0.45]{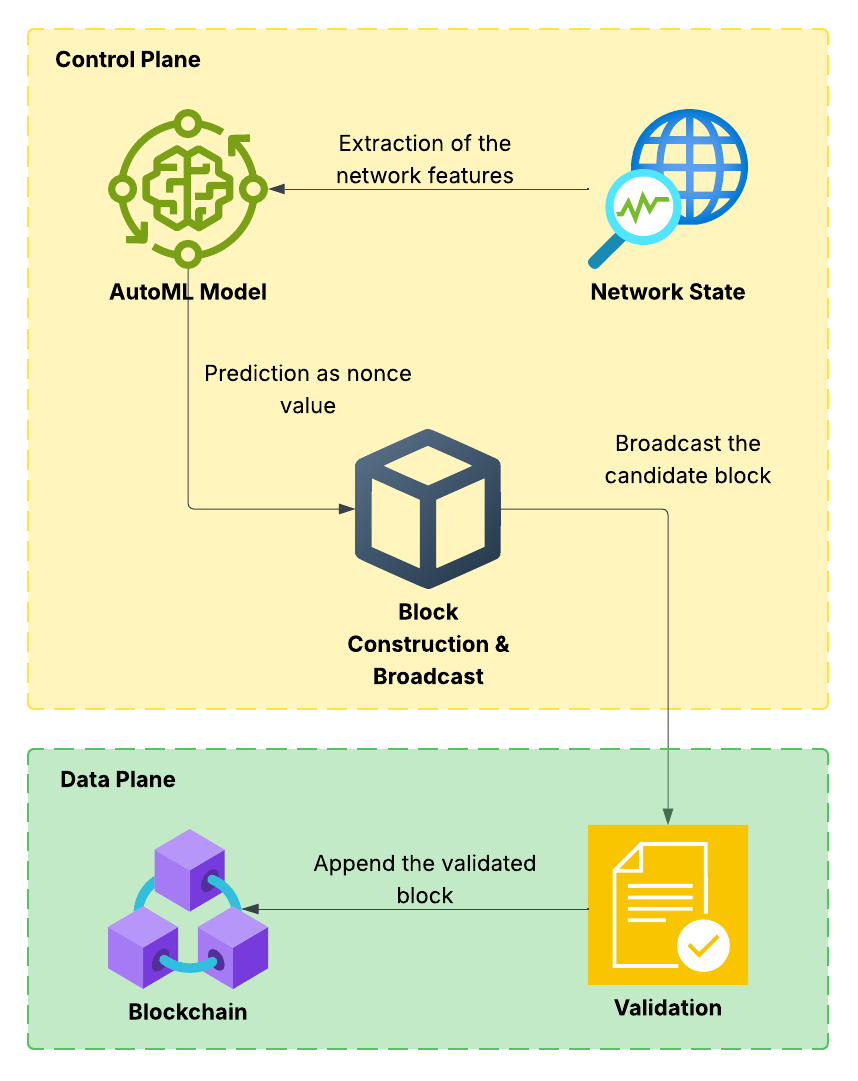}}
\vspace{-1em}
\caption{Proposed block mining method.}
\label{sys_block}
\vspace{-2em}
\end{figure}

\vspace{-0.5em}
To operationalize the proposed approach within the SDN-based blockchain infrastructure, we integrate AutoML-based regression models for dynamic and adaptive throughput estimation. Although nonce values in blockchain systems are conventionally random, our aim is to demonstrate that machine learning models can generate values with sufficiently random characteristics based on instantaneous network parameters. These values, while not used for deterministic prediction of nonce, serve as unique and unpredictable inputs that satisfy blockchain consensus mechanisms. The overall methodology is outlined in Fig. \ref{sys_block}, illustrating the interaction between the network layer, the AutoML model, and the blockchain layer.
\subsection{Model Objective and Dataset}
To develop a model capable of generating unique and random-like nonce values for secure block creation in post-disaster energy trading scenarios, we employed a dataset composed of measurable and easily obtainable network state parameters. Instead of relying on real-time feedback or deep network-layer integration, our objective is to train a model that can generate reliable outputs—nonce candidates—based on lightweight features such as delay, jitter, packet loss, and throughput. The dataset used in this study consists of 9000 samples, each representing a snapshot of a network state under varying conditions. These snapshots are structured as feature vectors composed of key Quality of Service (QoS) metrics, including:

\begin{itemize}
    \item \textbf{One-way delay (ms)}: The time taken for a packet to travel from source to destination.
    \item \textbf{Jitter (ms)}: The variation in packet delay.
    \item \textbf{Packet loss (\%)}: The ratio of packets lost during transmission.
    \item \textbf{Throughput (Mbps)}: The average rate of successful data delivery.
\end{itemize}

These parameters were selected due to their dual advantages:
(1) they are readily available in most network monitoring systems without requiring deep packet inspection or privileged access,
and (2) they are strongly influenced by real-world environmental and traffic conditions, which introduces a natural entropy into the dataset.
In this study, throughput is chosen as the primary target value—not because it is the final output to be used as a nonce, but because its variability across different network conditions makes it a suitable proxy to simulate randomness. The intuition is that, by learning to predict throughput from other QoS features, the trained model effectively captures the intrinsic noise and irregularity present in real network states. When used post-training, this model can generate unique, context-sensitive values that function as random-like nonce candidates for blockchain blocks.
This design allows us to decouple the model from the real-time constraints of the network and operate entirely based on sampled or previously measured data. This is especially critical in post-disaster scenarios, where connectivity may be limited or intermittent, and latency-sensitive computation must be minimized.
\vspace{-0.5em}
\subsection{Model Performances}
We trained and evaluated five different regression models using an AutoML pipeline to estimate the throughput (\textit{Mbps}) under disaster-aware SDN-based communication conditions. The models selected were chosen based on their performance and ability to generalize over diverse network metrics such as one-way delay, jitter, and packet loss.

\begin{table}
\centering
\caption{Performance Metrics of Evaluated Regression Models}
\label{tab:regression-results}
\begin{tabular}{lcccc}
\toprule
\textbf{Model} & \textbf{MAE} & \textbf{MSE} & \textbf{RMSE} & \textbf{R\textsuperscript{2}} \\
\midrule
Gradient Boosting Regressor & 0.418 & 0.385 & 0.620 & 0.619 \\
LightGBM Regressor          & 0.413 & 0.389 & 0.623 & 0.615 \\
Random Forest Regressor     & 0.417 & 0.395 & 0.628 & 0.609 \\
Extra Trees Regressor       & 0.426 & 0.413 & 0.642 & 0.591 \\
K-Neighbors Regressor       & 0.461 & 0.476 & 0.690 & 0.529 \\
\bottomrule
\end{tabular}
\vspace{-2em}
\end{table}

\begin{itemize}
  \item \textbf{MAE (Mean Absolute Error)}: Measures the average magnitude of the errors in a set of predictions, without considering their direction.
  \item \textbf{MSE (Mean Squared Error)}: Penalizes larger errors more significantly due to the squaring of the difference.
  \item \textbf{RMSE (Root Mean Squared Error)}: Provides a metric in the same unit as the output, which in this case is throughput (Mbps).
  \item \textbf{R\textsuperscript{2} Score (Coefficient of Determination)}: Indicates the proportion of variance in the dependent variable predictable from the independent variables.
\end{itemize}

Among the evaluated models, the Gradient Boosting Regressor achieved the best overall performance, with the lowest MAE and RMSE values, and the highest R\textsuperscript{2} score. 

\subsection{Randomness Evaluation}

Beyond predictive accuracy, an essential requirement in our methodology is the ability of a model to generate diverse and non-repetitive outputs. Since the ultimate goal is to obtain random-like values that can be used as valid nonce candidates, we introduce a metric called \textit{randomness rate}, calculated as:

\begin{equation}
R = \frac{U}{\sum_{i=1}^N P_i} \times 100
\end{equation}
Here, $R$ denotes the randomness rate, $P$ represents the number of unique predictions, $N$ is the total number of predictions, $U$ is the total number of unique predictions. $U$ can also be calculated as the total number of predictions ($\sum_{i=1}^N P_i$) minus the number of repeated predictions. This ratio provides an interpretable measure of how diverse the model outputs are. A randomness rate closer to $100\%$ indicates that the model rarely generates duplicate values, which is a desirable trait in nonce generation, particularly for maintaining the security and integrity of blockchain transactions in disaster scenarios.

In addition to randomness rate, we employ \textit{Shannon entropy} to quantify the unpredictability of the model outputs \cite{b16}. Shannon entropy is defined as:
\begin{equation}
H(X) = - \sum_{i=1}^{n} p(x_i) \log_2 p(x_i)
\end{equation}
where $p(x_i)$ is the probability of observing the prediction value $x_i$. Higher entropy values, normalized to the range $[0,1]$, indicate more uniformly distributed and less predictable outputs. Combining randomness rate and Shannon entropy allows for a comprehensive assessment of the model's ability to produce outputs suitable for secure and reliable nonce generation.

\subsection{Integration into SDN Control Plane}

The integration of the proposed AutoML-enhanced nonce prediction mechanism into the SDN control plane is essential for enabling intelligent and resilient blockchain-based energy trading in disaster scenarios. In our architecture, the SDN controller not only maintains a global view of the network topology and resource status, but also orchestrates the decision-making process for block creation and validation among base stations (BSs) acting as blockchain nodes.

During periods of limited connectivity and unstable infrastructure, the SDN controller leverages network telemetry—such as one-way delay, jitter, packet loss, and throughput—to extract the contextual state of the network. These parameters are collected in real-time and serve as input to a pre-trained AutoML model, which generates a pseudo-random but valid nonce to be embedded in the candidate block. This process eliminates the need for traditional Proof-of-Work (PoW) or deterministic nonce generation, thus significantly reducing the computational burden in energy-constrained disaster environments.

\begin{algorithm}[ht]
\caption{Proposed Proof of AutoML Algorithm.}
\label{alg:automl_nonce}
{\small
\begin{algorithmic}[1]
\State \textbf{Input:} EnergyTradingRequest, NetworkState
\State \textbf{Output:} ValidatedBlock with AutoML-predicted Nonce

\State Extract \textit{NetworkFeatures} from \textit{NetworkState}
\State $predicted\_nonce \gets \text{AutoML.predict}(NetworkFeatures)$
\State $encrypted\_tx \gets \text{Encrypt}(\text{TransactionList}, \text{BS\_private\_key})$
\State $block\_hash \gets \text{Hash}(predicted\_nonce,$ \newline
\hspace*{5.5em} $encrypted\_tx, prev\_hash, timestamp)$

\State Construct $CandidateBlock \gets \{encrypted\_tx,$ \newline
\hspace*{5.5em} $predicted\_nonce, block\_hash,$ \newline
\hspace*{5.5em} $prev\_hash, timestamp\}$

\State Broadcast $CandidateBlock$ to peer BS nodes for validation

\If{block is validated by consensus}
    \State Append $CandidateBlock$ to the blockchain ledger
\EndIf
\end{algorithmic}
}
\end{algorithm}

The overall block generation procedure is presented in Algorithm~\ref{alg:automl_nonce}. Upon receiving an \textit{EnergyTradingRequest}, the SDN controller initiates the block mining sequence. It first derives relevant network features from the monitored state, then queries the AutoML model to predict a nonce. The resulting transaction data are encrypted and hashed along with the predicted nonce, previous block hash, and timestamp to construct a new candidate block. This block is broadcast to peer BSs for consensus-based validation. If approved, it is appended to the distributed ledger.

By embedding this mechanism within the SDN controller, the system ensures that block creation remains feasible even when central authority or external connectivity is partially or fully disrupted. Furthermore, this architecture supports modular upgrades, such as adapting the AutoML model based on evolving network dynamics or incorporating more advanced randomness verification mechanisms. This tightly coupled integration enhances both the security and the sustainability of energy trading operations in crisis conditions.

\section{Performance Evaluation}

The performance evaluation experiments were conducted on a MacBook Air equipped with the Apple M2 chip and 16GB of RAM. The software environment utilized Python version 3.11.10 along with PyCaret version 3.32, a low-code machine learning library that automates the entire machine learning workflow, including tasks such as data preprocessing, model selection, and tuning. This setup enabled efficient and streamlined AutoML experiments, providing a reliable platform to measure and analyze performance metrics under realistic conditions. Table~\ref{tab:randomness} presents the randomness rates obtained from five different regression models used.

\begin{table}[h]
\centering
\caption{Results of Evaluated Models}
\label{tab:randomness}
\begin{tabular}{p{3.6cm}p{2cm}p{2cm}}
\toprule
\textbf{Model} & \textbf{Randomness Rate (\%)} & \textbf{Shannon Entropy (0–1)} \\
\midrule
Gradient Boosting Regressor & 97.6 & 0.9985 \\
LightGBM Regressor          & 99.9 & 0.9999 \\
Random Forest Regressor     & 100.0 & 1.0000 \\
Extra Trees Regressor       & 100.0 & 1.0000 \\
K-Nearest Neighbors Regressor & 98.8 & 0.9993 \\
\bottomrule
\end{tabular}
\vspace{-1em}
\end{table}

The results in Table~\ref{tab:randomness} reveal that both Random Forest and Extra Trees regressors achieved a perfect randomness rate of 100\% and a maximum Shannon entropy of 1.0000, indicating that they produced fully unique and highly unpredictable predictions within the test set. This outcome aligns with their intrinsic stochastic nature, which leverages bootstrapped sampling and random feature selection mechanisms during the training phase. 

On the other hand, while models such as Gradient Boosting and K-Nearest Neighbors did not reach perfect scores, their randomness rates (97.6\% and 98.8\%) and high Shannon entropy values (0.9985 and 0.9993) still reflect a substantial degree of diversity and unpredictability, sufficient for many nonce generation use cases. These insights reinforce the notion that ensemble methods, particularly those based on randomization, are promising candidates for random-like output modeling.
\vspace{-1em}
\begin{table}[h]
\caption{Model Training and Prediction Time Performance}
\centering
\begin{tabular}{lcc}
\toprule
\textbf{Model} & \textbf{Training (s)} & \textbf{Prediction (s)} \\
\midrule
Gradient Boosting Regressor & 0.143 & 0.042 \\
LightGBM Regressor & 0.219 & 0.043 \\
Random Forest Regressor & 0.282 & 0.071 \\
Extra Trees Regressor & 0.104 & 0.047 \\
K-Nearest Neighbors Regressor & 0.006 & 0.049 \\
\bottomrule
\end{tabular}
\label{tab:training_inference}
\end{table}

As shown in Table~\ref{tab:training_inference}, the K-Nearest Neighbors (KNN) Regressor demonstrated the fastest training time at just 0.006 seconds, owing to its lazy learning approach. However, its prediction time (0.049 seconds) is higher than some tree-based models, as KNN computes distances at inference. Among ensemble methods, the Extra Trees Regressor had the shortest training time (0.104 seconds) and maintained a low prediction time (0.047 seconds). Gradient Boosting and LightGBM Regressors required slightly longer training times (0.143 and 0.219 seconds, respectively) but achieved comparable prediction times around 0.04 seconds. The Random Forest Regressor showed the longest training (0.282 seconds) and prediction times (0.071 seconds) among the evaluated models. These results illustrate the trade-offs between training efficiency and inference latency across models. Nevertheless, since our objective is not to predict precise values but to generate random outputs, these timing metrics alone do not determine model suitability. Accordingly, the next subsection will analyze the randomness characteristics of each model.

The overall evaluation of the regression models reveals several important insights regarding their suitability for throughput prediction under dynamic and uncertain network conditions, such as those experienced during disasters. Tree-based ensemble methods, particularly \textit{Random Forest} and \textit{Extra Trees}, consistently demonstrated robust performance across accuracy, randomness rate, and Shannon entropy metrics.

While metrics such as MAE, MSE, RMSE, and $R^2$ provided insights into the models' predictive power, our goal extended beyond mere accuracy—toward generating values that exhibit high uniqueness and unpredictability, crucial for ensuring secure and non-repetitive nonce generation in blockchain-based energy transactions. In this regard, \textit{Random Forest Regressor} and \textit{Extra Trees Regressor} achieved a perfect randomness rate of 100\% and maximum Shannon entropy of 1.0000, confirming their ability to produce highly diverse and unpredictable outputs even when fed similar input features. By contrast, models like \textit{Gradient Boosting Regressor} and \textit{LightGBM Regressor} excelled in predictive accuracy but fell slightly short in randomness and entropy, with rates below 100\% and Shannon entropy slightly below 1.0.  Results indicate a trade-off between deterministic accuracy and stochastic variability that must be considered depending on the intended application context.

Overall, the findings suggest that when the primary objective is the generation of non-repetitive and unpredictable nonce values rather than highly accurate throughput estimation, randomness-focused ensemble models offer a promising solution. These insights not only validate our model selection strategy but also provide guidance for future implementations targeting secure energy trading infrastructures. Gradient Boosting and LightGBM models offered a strong trade-off between predictive accuracy and randomness/entropy. Ensemble-based models like Random Forest and Extra Trees achieved perfect randomness and maximum entropy but may be more computationally intensive during inference. All models produced nonce values that passed the randomness and unpredictability thresholds required for secure blockchain integration. 
The framework is robust and scalable, allowing deployment across multiple disaster-struck zones with decentralized control. These findings demonstrate that the proposed system not only achieves the desired cryptographic randomness and unpredictability but also adheres to the latency and reliability constraints of SDN-based blockchain networks under emergency conditions.
\vspace{-0.5em}
\section{Conclusion and Future Work}
\vspace{-0.5em}
In this study, we've proposes a secure energy trading framework for disaster scenarios using an SDN-assisted architecture with AutoML based randomness estimation. Rather than predicting a target variable, our focus has been on assessing models’ ability to generate unique, random-like nonce values from network parameters, supporting blockchain validation when central coordination fails. Using a 9000-row dataset of QoS metrics, five regression models were trained. Results show that ensemble methods, particularly Random Forest and Extra Trees, achieves strong accuracy and full randomness, making them suitable for nonce generation. Future work will integrate the mechanism into blockchain platforms for nonce validation, explore extensions such as energy forecasting, federated learning, and adaptive disaster recovery, and place greater emphasis on real-world disaster testbeds or simulations to demonstrate robustness beyond controlled conditions.
\vspace{-1em}
\section{Acknowledgment}
\vspace{-0.5em}
This paper has been supported by The Scientific and Technological Research Council of Turkey (TUBITAK) 1515 Frontier R\&D Laboratories Support Program for BTS Advanced AI Hub: BTS Autonomous Networks and Data Innovation Lab. Project 5239903.

\end{document}